\documentclass[12pt]{article}

\usepackage{UF_FRED_paper_style}
\usepackage{lipsum}  
\usepackage{graphicx}
\usepackage{inconsolata}

\usepackage{graphicx}
\usepackage{booktabs}
\usepackage{tabularx}
\usepackage{multirow}
\usepackage{graphicx}
\usepackage{subcaption}
\usepackage{pifont}
\usepackage{tabularx}
\usepackage[raggedrightboxes]{ragged2e}
\usepackage{fontawesome}

\usepackage{makecell, cellspace, caption}
\usepackage[table, dvipsnames]{xcolor}
\usepackage{array,booktabs}
\usepackage{subcaption}
\title{AI-generated Essays: Characteristics and Implications on Automated Scoring and Academic Integrity}

\author{Yang Zhong\\
    University of Pittsburgh \\
    {\texttt{yaz118@pitt.edu}}
\and  Jiangang Hao\\
ETS Research Institute 
\and Michael Fau\ss \\
   ETS Research Institute 
\and Chen Li \\
   ETS Research Institute 
\and Yuan Wang\\
   ETS Research Institute 
    }

\date{}

\begin{document}
\maketitle

\begin{abstract}

The rapid advancement of large language models (LLMs) has enabled the generation of coherent essays, making AI-assisted writing increasingly common in educational and professional settings. Using large-scale empirical data, we examine and benchmark the characteristics and quality of essays generated by popular LLMs and discuss their implications for two key components of writing assessments: automated scoring and academic integrity. Our findings highlight limitations in existing automated scoring systems, such as e-rater\textsuperscript{®}, when applied to essays generated or heavily influenced by AI, and identify areas for improvement, including the development of new features to capture deeper thinking and recalibrating feature weights. Despite growing concerns that the increasing variety of LLMs may undermine the feasibility of detecting AI-generated essays, our results show that detectors trained on essays generated from one model can often identify texts from others with high accuracy, suggesting that effective detection could remain manageable in practice.

\end{abstract}

\noindent \textbf{Keywords:} \textit{Automated Scoring, AI-generated Essay, Detection}


\section{Introduction}
Writing is a foundational literacy skill that is crucial for human communication and intellectual development. \citet{bazerman2008theories} highlights its indispensable role in enabling effective communication, and \citep{powell2012writing} (p. 12) even asserts that writing is ``the most important technology in the history of the human species, except how to make fire.'' Beyond its communicative function, writing is essential for fostering critical thinking, supporting learning across disciplines, and enabling individuals to organize and articulate complex ideas \citep{deane2008cognitive, adlerkassner2022writing,weigle2002assessing}. Given its significance, writing assessment plays a central role in evaluating language proficiency, communication skills, and analytical thinking. Standardized tests, such as the TOEFL (Test of English as a Foreign Language), IELTS (International English Language Testing System), and the GRE (Graduate Record Examination), include writing tasks designed to assess a test-taker's ability to construct coherent arguments, employ appropriate vocabulary, and demonstrate grammatical accuracy. 

The rapid advancement of large language models (LLMs) has significantly transformed the landscape of writing by enabling the generation of high-quality and coherent essays \citep{openai2024gpt4}.
As a result, AI-assisted writing has become increasingly prevalent in both educational and professional contexts. This widespread adoption of AI tools is reshaping how writing is created and assessed, with implications that span both the short and long term \citep{hao2024transforming}. In the short term, AI-generated essays raise serious concerns about academic integrity and test security. Students and test-takers may submit AI-assisted responses as their own work, challenging the ability of educators and testing organizations to evaluate writing skills and ensure fairness accurately. In the long term, the pervasive use of AI assistance could fundamentally alter human writing practices. Over-reliance on AI tools risks diminishing writing proficiency and critical thinking skills, as individuals may bypass the cognitive processes essential for effective communication and critical thinking. No doubt, these new developments carry significant implications for writing assessment.  

In this paper, we compare the characteristics and quality of AI-generated essays, examining two key implications for writing assessment: automated essay scoring and the detection of AI-generated responses for academic integrity. The study utilizes a large set of essays generated by various LLMs in response to prompts from the high-stakes GRE Analytical Writing Assessment (\url{https://www.ets.org/gre/test-takers/general-test/prepare/content/analytical-writing.html}) developed by Educational Testing Service (ETS) and leverage the official GRE scoring rubrics implemented by both well-trained human raters and the e-rater\textsuperscript{®} automated scoring engine \citep{Attali_Burstein_2006}. We specifically aim to address the following three research questions:

\begin{itemize}
    \item \textbf{RQ1}. How do essays generated by different large language models (LLMs) differ in their characteristics and quality as compared to human-written essays?
    \item \textbf{RQ2}. What are the limitations of current automated scoring engines when evaluating AI-generated essays compared to human raters, and how can these limitations be addressed?
    \item \textbf{RQ3}. How effective are detectors in identifying AI-generated essays using language and perplexity features when trained on essays produced by the same or different LLMs as those in the evaluation dataset?
\end{itemize}

To address \textbf{RQ1}, we begin by examining the linguistic and statistical characteristics of essays produced by a diverse set of high-performing LLMs as well as a random sample of human-written essays. This analysis includes evaluating adherence to prompt-specified length, language features, textual similarity, and perplexity\footnote{Perplexity is a measure of how well a language model predicts a sequence of words (tokens). Mathematically, it is the exponentiated average negative log-likelihood of the predicted tokens. Lower perplexity indicates that the model assigns a higher probability to the text, suggesting the text is more likely to have been generated by the language model.}. By comparing these characteristics across different models, we aim to uncover patterns that influence assessment outcomes, providing a nuanced understanding of how AI-generated texts differ from human-written responses. In response to \textbf{RQ2}, we assess the performance of a well-established automated essay scoring engine, e-rater\textsuperscript{®}, by comparing its scoring of AI-generated essays against that of human raters. This comparison allows us to identify specific limitations of the automated scoring system, from which we propose targeted improvements to better handle AI-generated or influenced essays. Finally, to explore \textbf{RQ3}, we investigate the effectiveness of detectors designed to identify AI-generated essays using language and perplexity features when training data and testing data are essays generated by the same or different LLMs.
Collectively, these insights advance our understanding of the implications of LLMs for writing assessments and offer practical strategies for addressing these emerging challenges.

\section{Literature Review}
\subsection{Advance in LLMs}
A language model in Natural Language Processing (NLP) refers to a statistical model that outputs a probability for a given sequence of texts \citep{jurafsky2008speech}. Over the past few years, with the advancement of GPU-based computing power, deep neural network-based modeling techniques, and the vast amount of data available on the internet, language models have grown significantly in size, leading to the emergence of so-called large language models. LLMs are characterized by their massive parameter counts, often ranging from billions to trillions, and are trained on extensive corpora containing diverse text sources, including books, websites, and publicly available research papers \citep{brown2020languagemodelsfewshotlearners, kaplan2020scalinglawsneurallanguage}.

Since the debut of ChatGPT, a variety of powerful LLMs have emerged, both open-source and proprietary. As of early 2025, the leading proprietary models include the GPT family from OpenAI \citep{openai}, the Gemini family from Google DeepMind \citep{google_deepmind_gemini}, the Grok family from xAI \citep{xai_grok}, and the Claude family from Anthropic \citep{anthropic_claude}. On the open-source front, several notable models have been introduced, including the Llama family from Meta \citep{meta_ai_llama}, the DeepSeek family \citep{deepseek}, the Mistral family \citep{mistral}, the Qwen family from Alibaba \citep{alibaba_qwen}, and the Phi family from Microsoft \citep{microsoft_phi}. For a comparison of their performance across tasks, readers may consult the LLM leader board \citep{llm_stats_leaderboard}. These models represent a diverse landscape of LLM capabilities, each contributing uniquely to advancements in natural language processing and AI-driven applications.

Among all the capabilities, a relevant one to this study is generating coherent and high-quality essays based on the user’s prompt. Currently, LLMs generate text using an autoregressive approach, which involves predicting and selecting each subsequent word or token sequentially based on the probability distributions learned from vast amounts of pretraining data \citep{chiang-lee-2023-large, openai2024gpt4, ouyang2022training, touvron2023llama}. This process begins with a prompt, which serves as the initial context, and the LLMs then generate one word at a time by evaluating the likelihood of all possible next words conditioned on the previous context, selecting the one with the highest probability. This step-by-step prediction continues iteratively, allowing the model to construct coherent and contextually appropriate sentences and paragraphs. Advanced techniques such as attention mechanisms and transformer architectures \citep{NIPS2017_3f5ee243} further enhance LLMs’ ability to capture long-range dependencies between words, ensuring that the generated text remains relevant and logically consistent with the initial prompt.

\subsection{Automated Scoring}
Automated scoring, which involves the use of computer algorithms to automatically evaluate the constructed responses, is a critical technology used in writing assessments. The initial conception of automated scoring dates back to the 1960s \citep{page1966imminence}  and the first automated scoring system was successfully implemented in the 1990s  \citep{page1995computer}. Several automated scoring engines have been used in operations for scoring short-constructed responses (e.g., Braun et al., 2006; \citeauthor{heilman-madnani-2015-impact}, 2015), long essays (e.g., \citeauthor{Attali_Burstein_2006}, 2006), and and mathematical equations \citep{fife2017mrater}.  Recently, the transformer architecture underlying most LLMs has shown excellent performance in automated scoring of short-constructed responses, essays, and speech ((Lottridge et al., 2022a; Lottridge et al., 2022b; Lottridge et al., 2022c).). 

Automated scoring models can be broadly categorized into two main types: prompt-specific vs. generic models \citep{flor2021text}. Each type serves distinct purposes and is optimized for different assessment contexts, providing valuable tools for evaluating written responses efficiently and accurately. Prompt-specific models are trained on responses to a particular prompt or a closely related set of prompts. These models are fine-tuned to capture the specific expectations, vocabulary, and content relevant to a given prompt. By focusing on prompt-specific features (such as n-grams or semantic embeddings), they offer more accurate and context-sensitive scoring, ensuring that responses are evaluated against the precise criteria expected for that task. In practice, prompt-specific models are typically employed for scoring short constructed responses to open-ended questions.

In contrast, generic models are designed to operate across a wide range of topics and prompts. Unlike prompt-specific models, they focus on assessing broader constructs that are not tied to particular contexts, such as writing competency, analytical reasoning, or critical thinking. Generic models typically rely on non-contextualized features, such as grammar, organization, and logical structure, rather than features confined to specific contexts. As a result, they can effectively evaluate responses to new or unfamiliar prompts, offering greater flexibility across assessment settings. In practice, generic models are widely used for scoring longer essays in standardized testing. For example, e-rater\textsuperscript{®} is a generic automated scoring model that has been used to score essays in assessments such as TOEFL and GRE. In this study, we focus on the generic model. Unless otherwise specified, all subsequent references to automated scoring refer to this type of model.

Before the advent of LLMs, automated scoring systems typically relied on a two-step process: first, extracting numerical features from texts using natural language processing (NLP) techniques, and second, mapping these features to scores through statistical or machine learning models \citep{Attali_Burstein_2006, chen2016buildingerater}. However, since the introduction of LLMs, two new approaches have emerged. The first approach is end-to-end, where texts and their corresponding scores are used to fine-tune LLMs (such as BERT), establishing a direct mapping between texts and scores without the need for explicit feature extraction. The second approach involves directly instructing LLMs to score texts by providing them with scoring rubrics and a few examples \citep{hao_cui_kyllonen_kerzabi_liu_flor_2024}, which heavily relies on the in-context (few-shot) learning capabilities of LLMs, enabling them to generate scores directly based on the provided guidelines. While LLM-based approaches often outperform traditional methods in terms of accuracy, they also introduce new challenges. Specifically, the end-to-end and rubric-based approaches significantly reduce the explainability of the models, making it more difficult to understand the basis for the assigned scores \citep{Casabianca2025Validity}. Additionally, these methods can amplify biases present in the training data, leading to bias across demographic groups \citep{johnson2024examining}.

\subsection{Detection of AI-generated Essays}
The detection of machine-generated essays initially focused on identifying texts specifically designed to deceive automated scoring systems. For instance, the Babel essay generation system \citep{babel2014generator} was created to generate essays intended to mislead these scoring engines. This tool accepts a list of keywords as input and uses them to construct essays that exhibit well-formed syntactic structures and advanced vocabulary. Although these essays lack genuine semantic depth, they appear linguistically complex and coherent, making it difficult for automated systems to recognize them as artificially generated. In response, \citet{cahill2018developing} proposed an advisory flagging system capable of detecting such machine-generated essays by identifying patterns indicative of artificial construction. 

With the advent of LLMs, the quality of generated texts has improved significantly, prompting the development of more advanced detection systems. For instance, \citet{Yan2023-ha} demonstrated that essays generated by GPT-3 in response to large-scale writing tests could be detected with high accuracy using two distinct approaches: a language feature-based supervised learning method and an end-to-end classifier built on a fine-tuned LLM, RoBERTa \citep{liu2019roberta}. Following the release of ChatGPT in November 2022, detecting AI-generated texts quickly became a focal point of \citep{Crothers2023survey, haoandfauss, tian2023gptzero}, and and several detectors have been developed to identify ChatGPT-generated essays in large-scale writing assessments \citep{haoandfauss, jiang2024detecting, jiang2024towards}. These detectors leverage a combination of language and perplexity-based features, achieving detection accuracies exceeding 98\%. Additionally, detectors based on keystroke dynamics have also been introduced, proving effective at distinguishing authentic draft writing and copywriting by analyzing typing patterns and temporal characteristics  \citep{deane2025keystroke, jiangzhang2024keystroke}. A lingering question, however, is whether models trained on data from one LLM can reliably detect essays generated by other LLMs, especially as new models continue to emerge every few months.

\section{Methods}

\subsection{GRE Analytical Writing and E-rater\textsuperscript{®} Automated Scoring Engine}
The GRE analytical writing is well known for assessing test taker’s ability to think critically and communicate complex ideas effectively. It consists of a 30-minute writing task,” analyze an issue,” which requires examinees to ponder a topic, evaluate its intricacies, and formulate a well-reasoned argument buttressed by examples and explanations. The performance is scored into six levels: fundamentally deficient (1), seriously flawed (2), limited (3), adequate (4), strong (5), and outstanding (6). More details of the scoring framework can be found in Appendix A. Overall, the assessment aims to evaluate a candidate’s ability to articulate and substantiate complex thoughts, construct and evaluate arguments, and maintain a coherent and focused discourse throughout the essay (ETS, n.d.-a).

Each essay in the GRE analytical writing is scored by a well-trained human rater and by ETS’s automated scoring engine, e-rater®. Human raters undergo a robust training and certification process: they must meet strict eligibility criteria, complete training on the scoring rubric, and pass a calibration test at the start of each scoring session to ensure scoring accuracy. During scoring, pre-scored “validation” essays are interspersed to monitor accuracy in real time; if a rater fails to maintain consistency even after retraining, they are dismissed \cite{wendler2019Rater}. The e-rater\textsuperscript{®} takes in an essay and outputs a score (in the range of 1 to 6), which is a weighted sum of a set of language features, such as grammar, usage, mechanics, styles, organization, development, and word complexity \citep{chen2016buildingerater}. Table \ref{tab:features} gives a summary of a subset of language features we focused on in this paper. These features are computed using NLP techniques, and their weights in the final score are determined by linear regression or machine learning models using a large corpus of essays scored by human raters. As such, the final e-rater\textsuperscript{®} score depends on both the selection of language features and their corresponding weights based on human-scored essays. Extensive studies and literature documents that the e-rater score is well agreed with human scores \cite{ets_gre_compendium,ets_erater_about, williamson2012Framework}.

\begin{table*}[h!]
\centering
\footnotesize

\begin{tabular}{p{1.1in}p{4.7in}}
 \toprule

\textbf{Feature} & \textbf{Description} \\
\midrule
\textbf{\multirow{2}{*}{Grammar}}& \RaggedRight{A statistic based on the number of errors related to pronouns, run-ons, missing possessives, etc. The more negative the value, the more errors.} 
     \\ 
     \midrule
\textbf{\multirow{2}{*}{Mechanics}}& \RaggedRight{A statistic based on the number of errors related to capitalization, punctuation, commas, hyphens, etc. The more negative the value, the more errors.} 
     \\ 
     \midrule
\textbf{\multirow{2}{*}{Style}}& \RaggedRight{A statistic based on the number of errors related to repetition of words, inappropriate words, etc. The more negative the value, the more errors.} 
     \\ 
     \midrule
    \textbf{\multirow{2}{*}{Usage}}& \RaggedRight{A statistic based on the number of errors related to missing/wrong articles, nonstandard verbs, etc. The more negative the value, the more errors.} 
     \\ 
     \midrule
      \textbf{\multirow{1}{*}{Organization}}& \RaggedRight{The logarithm of the number of discourse units} 
     \\ 
     \midrule
 \textbf{\multirow{1}{*}{Development}}& \RaggedRight{The logarithm of the mean length of discourse units} 
     \\ 
     \midrule

\textbf{\RaggedRight{Word Complexity}}& \RaggedRight{The average length of words} \\


\bottomrule
\end{tabular}
\caption{Language features from e-rater\textsuperscript{®} engine output used in this study.}\label{tab:features}
\end{table*}

\subsection{LLM and AI Essay Generation}
Since the release of ChatGPT in November 2022, LLMs have undergone an unprecedented period of rapid advancement, with increasingly powerful models emerging every few months. This study focuses on a range of LLMs, both proprietary and open-source, available between February 2023 and May 2024, a period that captures the peak of this rapid development. After this point, progress in LLMs has shifted from scaling data and model size toward enhancing reasoning capabilities, primarily through more sophisticated internal prompting techniques rather than additional training on new datasets. Within the proprietary sector, we chose the top ones, including GPT-4o from OpenAI and Gemini1.5 from Google. These closed-source models represent the forefront of AI research in the industry, offering a glimpse into the state-of-the-art capabilities in language generation. In addition to these new models, we also included some older ones, such as GPT-4 and GPT-3.5 Turbo from OpenAI and Bard from Google, which were introduced in the middle of 2023.

On the other hand, we also included some leading open-source LLMs in our evaluation. Specifically, we considered the recent Llama-3 8B model, which has been the subject of extensive research since its release \citep{dubey2024llama3herdmodels}. In addition, we experimented with some older models from the middle of 2023, including the Llama-1 13B \citep{touvron2023llama}, as well as its two finetuned variants: Vicuna \citep{vicuna2023} and Koala \citep{koala_blogpost_2023}, both of which underwent specialized optimization processes to further refine their language generation mechanisms. The 13-billion parameter version was selected for each of these earlier models. In addition to the Llama family of models from Meta, we included another popular LLM, Mistral-7B-Instruct-v0.1, from Mistral AI, which is comparable in size to the Llama-3 model. The detailed version information of these models is presented in Appendix \ref{appendix:llm_detail}.

All writing prompts of the GRE writing assessment are publicly available (ETS, n.d.-b). To alleviate the risk of biases introduced by an individual writing prompt, we randomly selected two writing prompts for this study. To prompt LLMs to generate the essays, we sent the original GRE writing prompt plus an instruction on length: “Please keep the response to about 500 words.”, where the 500 words is the typical length of GRE essays. The resulting prompts are listed below.

\begin{itemize}
    \item \textbf{Prompt 1}: Governments should not fund any scientific research whose consequences are unclear. Write a response in which you discuss the extent to which you agree or disagree with the recommendation and explain your reasoning for the position you take. In developing and supporting your position, describe specific circumstances in which adopting the recommendation would or would not be advantageous and explain how these examples shape your position. Please keep the response to about 500 words.

    \item \textbf{Prompt 2}: To understand the most important characteristics of a society, one must study its major cities. Write a response in which you discuss the extent to which you agree or disagree with the statement and explain your reasoning for the position you take. In developing and supporting your position, you should consider ways in which the statement might or might not hold true and explain how these considerations shape your position. Please keep the response to about 500 words
\end{itemize}

We generated 100 essays for each prompt from each LLM, leading to a total of 2,000 essays from the ten models. Each LLM comes with a set of hyperparameters that can be adjusted to control the text generation. Among these hyperparameters, temperature controls the randomness of token selection by scaling the logits before applying the SoftMax function. Given logits $z_i$, the probability of selecting token i with temperature T is \( p_i = \frac{\exp(z_i / T)}{\sum_j \exp(z_j / T)} \). Lower temperatures make the output more deterministic by sharpening the distribution while higher temperatures increase the randomness, leading to more diverse but potentially less coherent texts. 

For comparison purposes, we set the temperature to 0.4 (the default value used in earlier versions of the web-based ChatGPT to strike a balance between diversity and quality) for each model, except for Bard, where this parameter configuration was not accessible. Here, we caution that different LLMs may use varying scaling for their temperature, meaning that the same numerical value may affect the randomness and diversity of responses differently across models. In addition to AI-generated essays, we randomly sampled 100 human-written essays for each of the two prompts from the 2021 GRE writing test, a period before ChatGPT-like AI systems were available. This ensures that the selected essays were authentically human-written. We chose 100 human essays to align with the number of AI-generated essays per prompt per LLM, facilitating direct comparison. The decision to generate 100 AI essays per prompt per LLM was made to strike a balance between capturing sufficient variability in performance and maintaining feasibility for analysis. Together, these human- and AI-written essays provide a comparable and well-controlled basis for evaluation as well as training data for developing detection models.  

\subsection{Essay Characteristics}
To address \textbf{RQ1} (how essays generated by different large language models (LLMs) differ in their characteristics and quality compared to human-written essays), we examine a set of selected features of the AI-generated essays. These characteristics were chosen because they are highly relevant to the other two research questions, e.g., automated scoring and the detection of AI-generated essays.

\textit{Essay Length.}  We compute the length of the generated essay to check how well each LLM adheres to the 500 words in the prompt. 

\textit{Essay Similarity.} We compute the pairwise similarity between essays generated by the same LLM to assess the variation in the LLM’s outputs under the same prompt. There are two commonly known types of similarities between texts: verbatim similarity, which describes how closely two texts match in exact wording or surface structure, and semantic similarity, which shows how similar two texts are in meaning, regardless of specific word choices. Comparing these two types of similarities for pairwise essays generated by LLMs under the same prompt could reveal how well the LLMs maintain meaning across different expressions of the same content, having important implications for automated scoring and the detection of AI-generated essays. 

For semantic similarity, we first convert each essay into a document vector using a popular neural embedding model \textit{jina-embedding-v3} model \citep{jina-v3}. This embedding model was chosen based on its high performance on text similarity benchmarks of the MTEB English leaderboard and its superior ability to handle long documents. Once the essays are converted to vectors, we compute the cosine similarity between pairs of vectors corresponding to essays generated from the same LLM under the same prompt. Cosine similarity measures the similarity between two vectors by comparing the angle between them, regardless of their magnitude. It ranges from –1 to 1, where values close to 1 indicate strong similarity (vectors pointing in the same direction), values near 0 suggest little to no similarity (orthogonal vectors), and values near –1 reflect strong dissimilarity or opposite direction. In practice, for text embeddings, cosine similarity above 0.8 typically indicates high similarity, around 0.5 suggests moderate similarity, and below 0.2 indicates minimal similarity. 

For verbatim similarity, we represent each essay as a trigram vector, which encodes the frequencies of overlapping three-word sequences in the text. This representation is effective in capturing textual similarities for test security applications \citep{choi2024autoesd}. Based on the trigram vectors, we computed pairwise cosine similarities for each pair of essays generated by the same LLM for the same prompt. 

\textit{Language Features.} We compute the language features as shown in Table 1 for each essay using e-rater\textsuperscript{®} and then compare the statistics of these features across the essays generated by different LLMs. 

\textit{Perplexity Distribution.}  Perplexity characterizes the likelihood of a sequence of words (in an essay) being generated from a given language model and plays a crucial role in detecting AI-generated texts. Most LLMs used in this study do not allow the direct computation of perplexity. Therefore, we selected GPT-2 as a common language model to calculate the perplexity of essays generated by all LLMs. As such, the perplexity here is a relative measure of how likely the essays have been generated by GPT-2. Prior research has demonstrated that perplexity, computed based on a common language model, is highly informative for detecting AI-generated essays. (e.g. \citeauthor{haoandfauss}, 2024, \citeauthor{jiang2024detecting}, 2024a).

\subsection{Essay Scores}
We recruited expert human raters from the GRE program to score the 200 essays (100 from each of the two writing prompts) generated by each LLM and humans. The raters applied the same scoring rubric and procedures used in the operational GRE program, assigning each essay a score on a scale from 1 to 6 (see Appendix \ref{appendix:complete_table1}  for score definitions). The source of the essay, i.e., written by humans or AI, has been deliberately hidden from the human raters, so they were unaware the essays were generated by AI. In addition to human scores, we used the e-rater\textsuperscript{®} engine to score each AI-generated essay and compared its scores with those assigned by human raters. The discrepancies between the two sets of scores highlight the limitations of the current automated scoring system in evaluating AI-generated content. They also highlight opportunities for improvement, helping to better prepare automated scoring engines for the growing prevalence of AI-generated and AI-assisted texts, which is exactly the focus of \textbf{RQ2}.

\subsection{Detection of AI-generated Essays}
As the quality of essays generated by LLMs continues to improve, concerns are rising in education about students submitting AI-generated assignments, which poses a significant challenge to academic integrity. These concerns have become a significant risk, even in high-stakes testing, which began with the rise of remotely proctored exams following the COVID-19 pandemic \citep{haoandfauss}. The detection of AI-generated essays in large-scale assessments has been a widely discussed topic \cite{haoandfauss, jiang2024detecting, Yan2023-ha}, with most previous work focusing on detecting essays from a single LLM. However, with new and more powerful LLMs emerging every few months, it becomes increasingly difficult to identify which LLMs are being used. This highlights the importance of examining the detectability of AI-generated essays across multiple LLMs, which is the \textbf{RQ 3} of this study. 

Specifically, we consider two scenarios: within-model detectability and cross-model detectability. In the within-model scenario, we assess how well classifiers trained on essays generated by a specific LLM can detect essays produced by that same LLM. In the cross-model scenario, we examine how effectively a classifier trained on data from one LLM can detect essays generated by other LLMs. In both cases, the classifiers were developed using the seven language features in Table 3 and perplexity-based features (including the whole essay perplexity score, the mean, median, maximum, minimum, and 10, 20, 30, ..., 90 percentiles of the sentence perplexity in each essay), respectively.  We experimented with multiple popular classifiers, including XGBoost \citep{chen2016xgboost}, Random Forest \citep{breiman2001random}, Gradient Boosted Machine \citep{friedman2001greedy} and Support Vector Machine \citep{cortes1995support}
. We found that XGBoost consistently yielded the best results. So, we only show the results of the XGBoost classifier.

\section{Results}

\subsection{Comparing Characteristics of Essays}
\begin{figure}[h!]
    \centering
    \includegraphics[width=0.8\columnwidth]{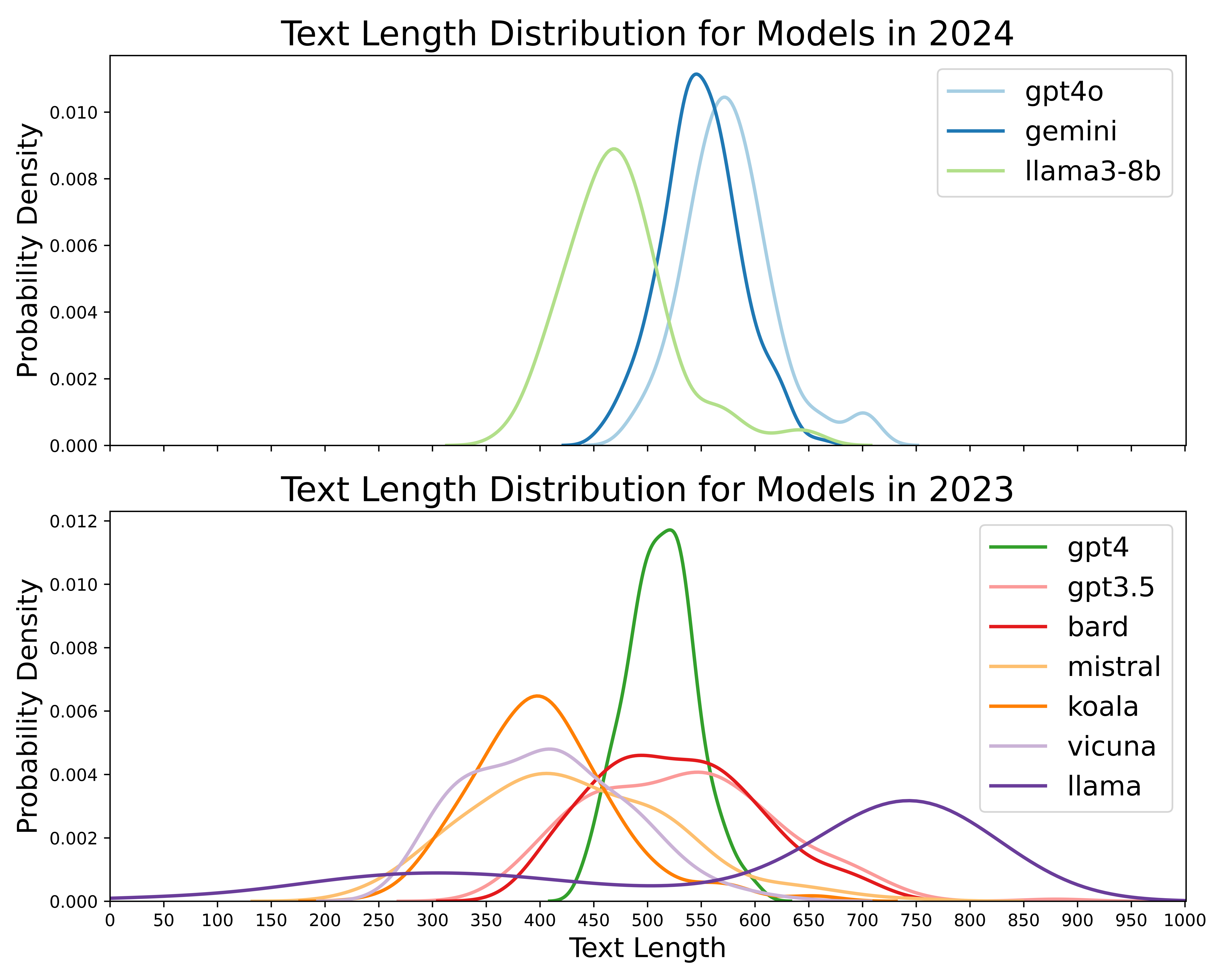}
    \caption{Distribution of the length of essays generated by different LLMs.}
    \label{fig:length_fig}
\end{figure}

\hspace*{2em}\textit{Essay Length Adherence.} We introduced a length requirement, i.e., about 500 words, when prompting LLMs to generate essays. However, not all LLMs follow the instructions strictly. Figure \ref{fig:length_fig} shows the distribution of the lengths of the generated essays in terms of the number of words. Overall, we observe that newer models exhibited smaller variances and better alignment with the 500-word instruction. For more recent models in 2024, GPT-4o and Gemini generated essays with lengths slightly greater than 500, while Llama3-8b generated an average of 460 words. For older models, close-sourced models performed better than open-sourced models (Mistral, Vicuna, Koala, and Llama) in following the length instruction. Essays generated by Llama displayed the largest length variability, indicating that the model struggled to consistently produce long and high-quality essays, which echoed the findings on essay quality.

\begin{figure}[h!]
    \centering
    \includegraphics[width=0.85\columnwidth]{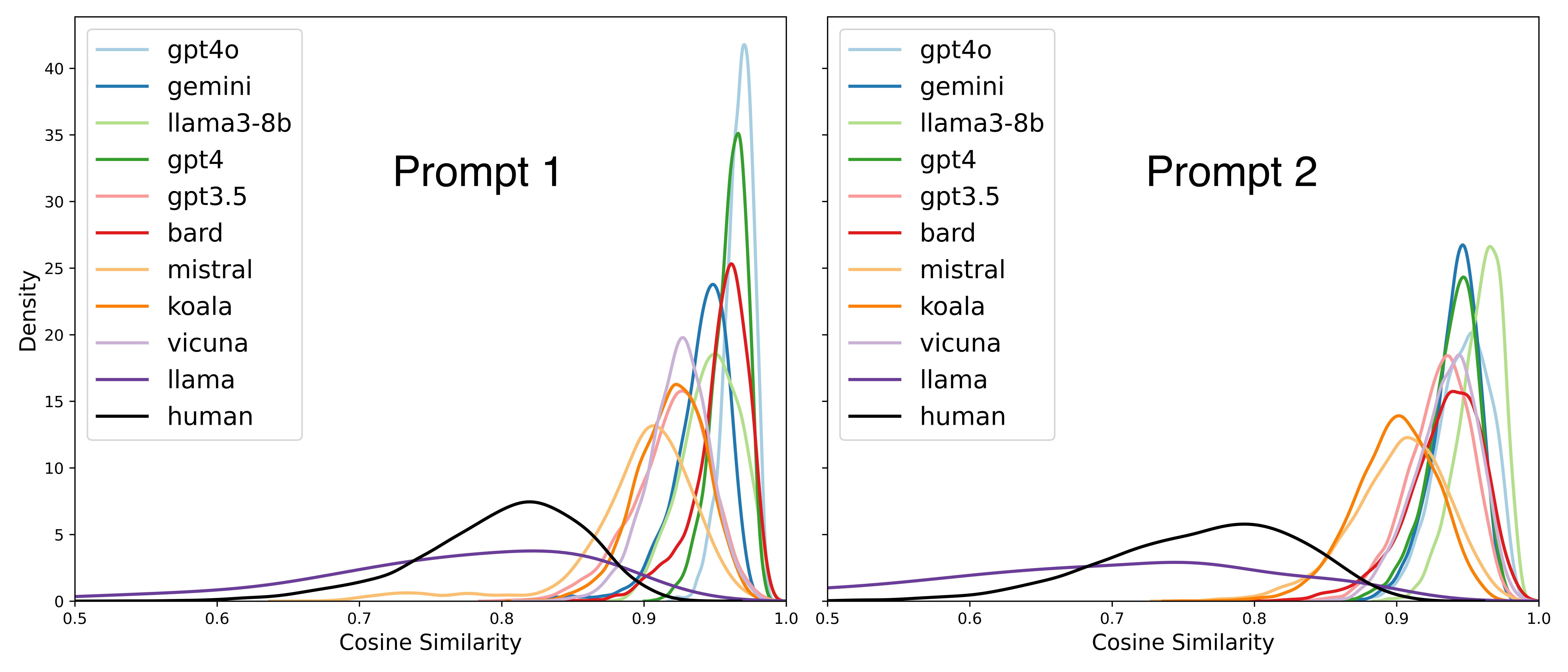}
    \caption{Semantic similarity between pairs of essays generated by the same LLM and prompt.}
    \label{fig:similarity}
\end{figure}

\textit{Essay Similarity.} For the pair-wise semantic similarity of essays generated by different LLMs under the same prompt, the results are shown in Figure \ref{fig:similarity}. 

\begin{figure}[t!]
    \centering
    \includegraphics[width=0.87\columnwidth]{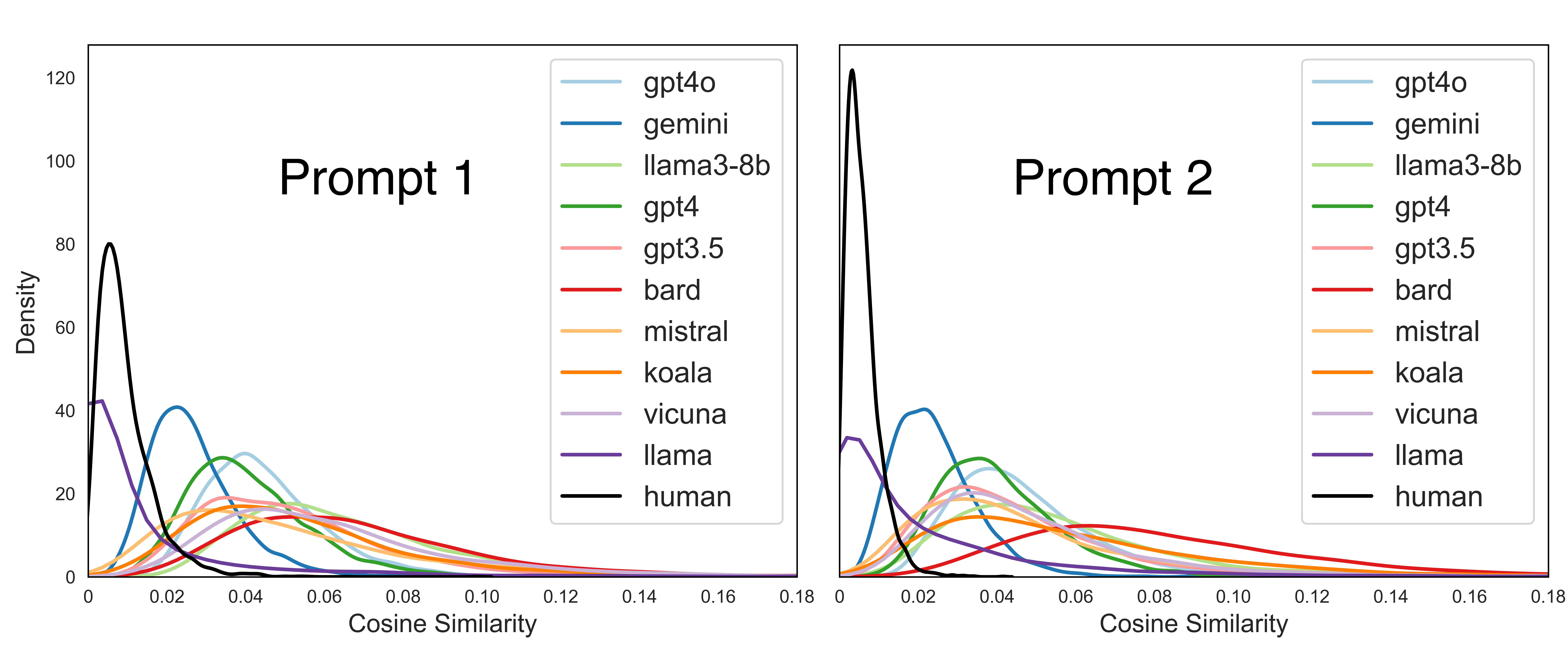}
    \caption{Verbatim similarity between pairs of essays generated by the same LLM and prompt.}
    \label{fig:trigram_similarity}
\end{figure}

We observed that essays generated by the Llama model displayed the least similarity, comparable to human-written essays. Models such as GPT-4, Gemini, and GPT-4o tend to have distributions with higher mean cosine similarity, indicating that their generated essays are more semantically similar to each other. The distributions between the two prompts differ slightly, with some models demonstrating higher density and peaks for one prompt than another (i.e., GPT-4o has a higher density for prompt 1, while Llama3-8b obtains the highest mean similarity for prompt 2).
For the verbatim similarity, the results are shown in Figure \ref{fig:trigram_similarity}.  
Overall, human essays displayed the lowest similarity, while different LLMs showed varying levels of similarity in their generated texts. Despite the essays from the same LLM and prompt showing a high level of semantic similarity, the verbatim similarity is much lower, though they are generally higher than that from human-written essays.
\begin{table*}[t!]
\centering
\footnotesize
\setlength{\tabcolsep}{1pt}
\newcommand*\rot[1]{\rotatebox{90}{#1}}
\begin{tabular}{clccccccc}
\toprule
& &	\textbf{Grammar}$\uparrow$&\textbf{Mechanics}$\uparrow$	& \textbf{Usage}$\uparrow$	& \textbf{Style}$\uparrow$ &	\textbf{Organization}& \textbf{Development}	& \textbf{Word-Complexity} \\ 
\midrule
\rowcolor{lightgray}\multicolumn{9}{c}{\textit{mid-2024}}\\
\midrule
 & \textbf{GPT-4o} & -0.075±0.001	& -0.050±0.002&	-0.035±0.002	&-0.328±0.002	&2.457±0.007	&3.794±0.008	&5.895±0.014\\
&\textbf{Gemini}	& -0.080±0.001&	-0.048±0.002&	-0.059±0.002	&-0.317±0.002&	2.344±0.008&	3.824±0.008&	6.100±0.018\\
&\textbf{Llama3-8b}	& -0.068±0.001	& -0.014±0.002	& -0.044±0.002	&-0.392±0.002	&2.233±0.008&	3.836±0.007	& 5.547±0.015 \\
\midrule
\rowcolor{lightgray}\multicolumn{9}{c}{\textit{mid-2023}}\\
\midrule
 & \textbf{GPT-4} &	-0.075±0.001&	-0.022±0.002	&-0.048±0.002	&-0.356±0.002	&2.359±0.007	&3.762±0.007	&5.610±0.018 \\
&\textbf{GPT-3.5}	 &-0.072±0.001 &	-0.013±0.001 &	-0.031±0.002	 &-0.378±0.002	 &2.343±0.011	 &3.865±0.010	 &5.339±0.015 \\

&\textbf{Bard}	 &-0.082±0.001	 &-0.015±0.001	 &-0.038±0.002	 &-0.384±0.002	 &2.552±0.013	 &3.602±0.012	 &5.023±0.016 \\

&\textbf{Mistral} &	-0.082±0.001&	-0.019±0.002	&-0.037±0.002	&-0.417±0.003	&2.292±0.016&	3.704±0.014	&5.335±0.017\\
&\textbf{Koala}	&-0.064±0.001&	-0.010±0.001	&-0.045±0.002	&-0.420±0.002&	2.208±0.011&	3.773±0.010	&5.147±0.017 \\
&\textbf{Vicuna} &	-0.067±0.001	& -0.011±0.001	&-0.034±0.002&	-0.394±0.002	&2.198±0.009	&3.759±0.011	&5.238±0.016 \\
&\textbf{Llama}	& -0.081±0.001 &	-0.036±0.004	&-0.041±0.003&	-0.344±0.008&	1.830±0.046&	4.428±0.052	&4.648±0.037 \\

\midrule
& \textbf{Human}	& -0.082±0.001	& -0.103±0.002	& -0.076±0.002	&-0.333±0.002&	2.114±0.012 &	4.290±0.013	& 5.111±0.015 \\

\midrule
\end{tabular}
\caption{Mean and standard error of the mean corresponding to the key language features from different LLMs. $\uparrow$ means the negative values close to 0 are more favored.}\label{tab:feature_stats}
\end{table*}

\begin{figure}[h!]
    \centering
    \includegraphics[width=0.8\columnwidth]{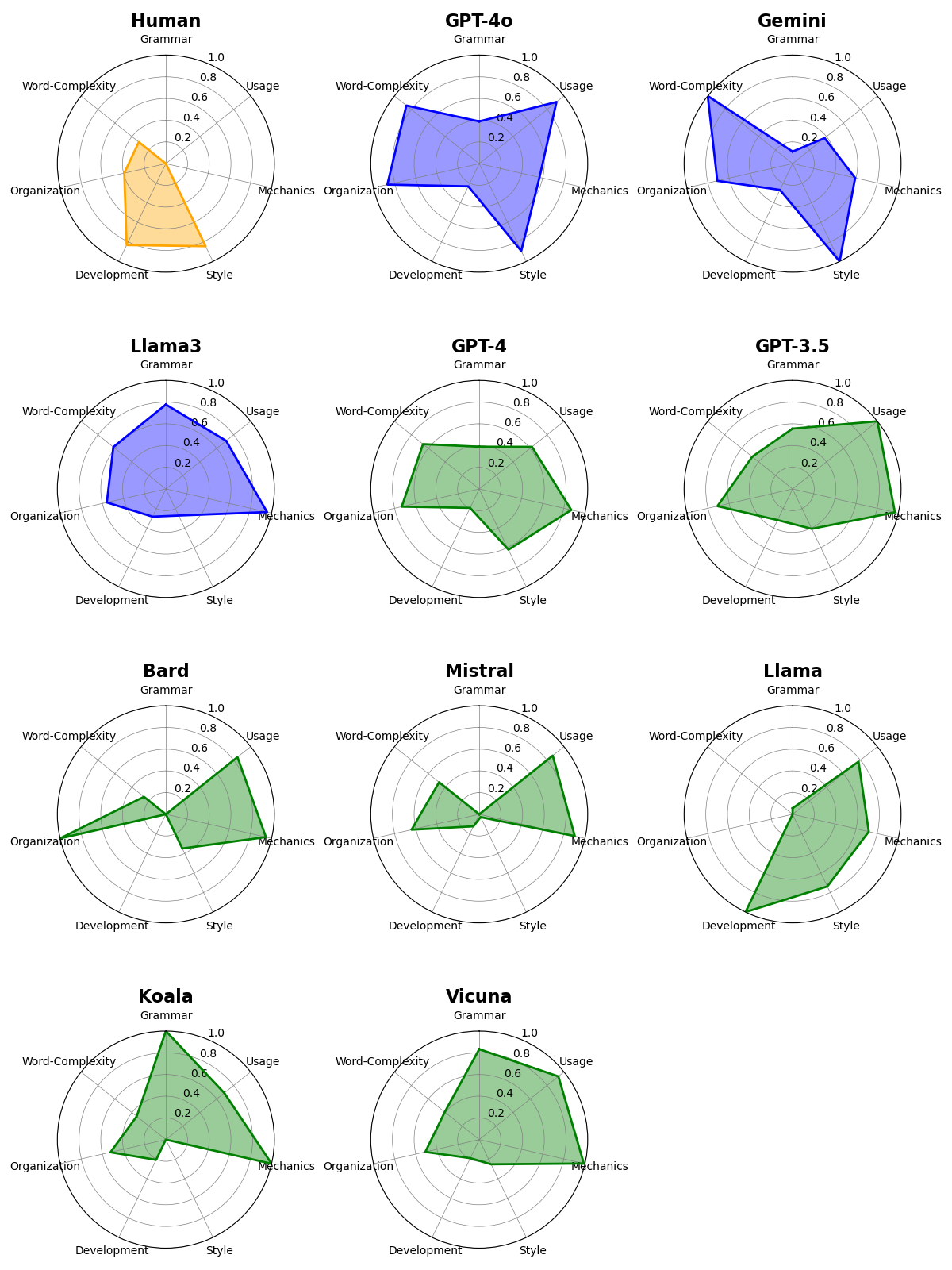}
    \caption{Comparison of the language features of the essays from different LLMs and humans. The ones in blue colors are models in 2024 and the ones in green colors are those from 2023. The original values of each feature from multiple LLMs have been rescaled using MinMax scaling, which transforms each value to a range between 0 and 1 based on the minimum and maximum observed values by \( x' = \frac{x - \min(x)}{\max(x) - \min(x)} \). After the scaling, higher values indicate better essay quality in that dimension.}
    \label{fig:radar_language}
\end{figure}

\textit{Language Features.} The language features from e-rater\textsuperscript{®} allow us to compare more detailed aspects of the generated essays across the LLMs. Table \ref{tab:feature_stats} shows the mean and standard error of the language features. It is important to note that grammar, mechanics, usage, and style features are negative by convention, as they represent errors: the closer the value is to 0, the fewer the errors. In contrast, organization, development, and word complexity features are positive, with higher values indicating better performance in these areas. To visualize the differences across AI-generated and human-written essays, we apply MinMax scaling to normalize all features to the same range (from 0 to 1) and present them in Figure \ref{fig:radar_language}. Based on the results, we have the following observations.

Humans demonstrate stronger performance in \textit{development} but show weaker performance in \textit{grammar, mechanics, and usage}. This suggests that human writing is strong in terms of structure and coherence but prone to technical errors compared to LLMs. The GPT family (particularly GPT-4o and GPT-4) shows strong performance in grammar, mechanics, and usage, with wide coverage on the radar plots, indicating an all-around high performance with fewer errors and more fluent text generation. The 2024 models generally perform better than the 2023 models, suggesting steady progress in these LLMs.

\begin{figure*}
    \centering
    \includegraphics[width=1\columnwidth]{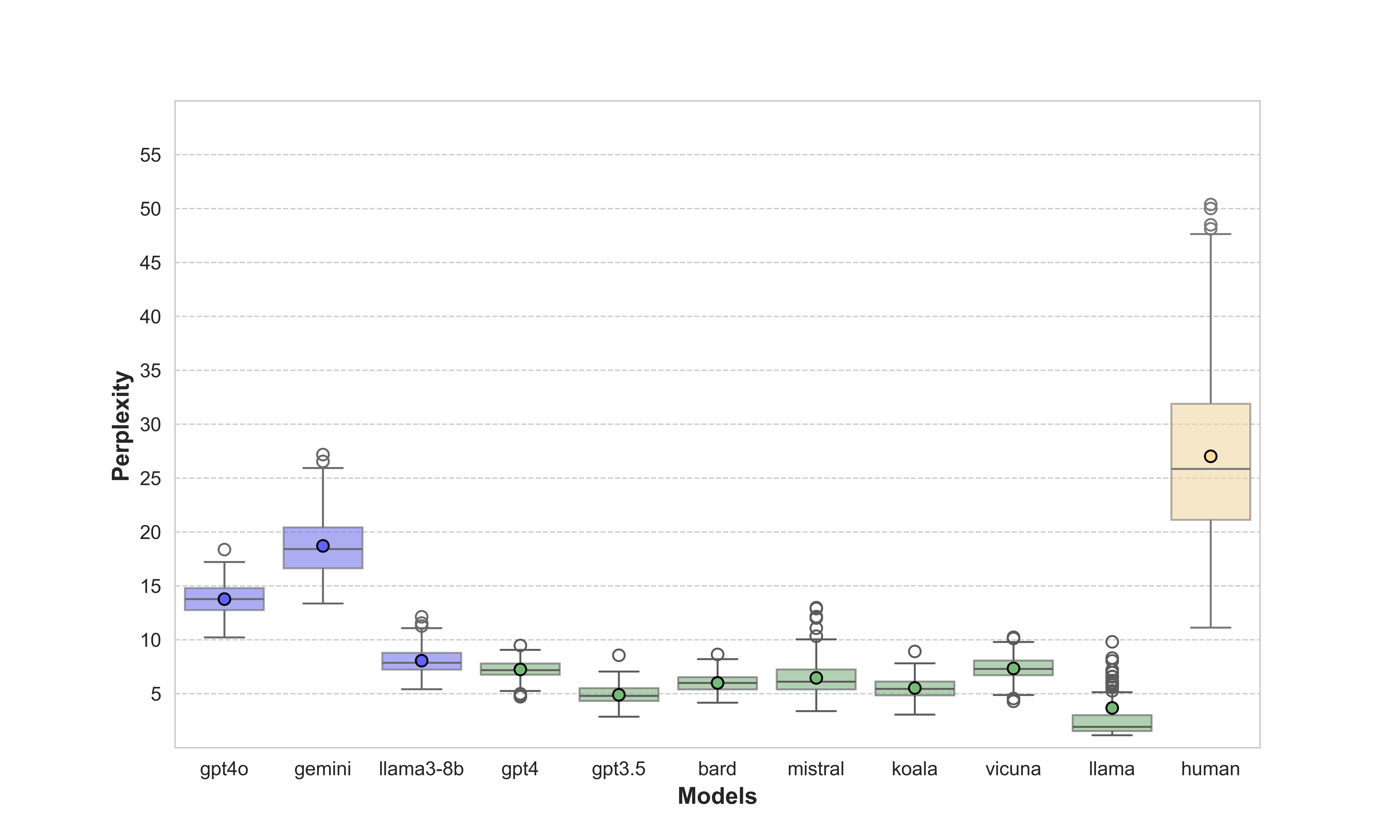}
    \caption{Boxplot of the essay perplexity distribution for different models. The ones in blue colors are from essays generated by 2024 LLMs and the ones in green colors are those from 2023 LLMs. }
    \label{fig:perplexity}
\end{figure*}

\textit{Perplexity.} Figure \ref{fig:perplexity} compares the distribution of perplexity computed from essays generated by LLMs and written by humans. The perplexity of human-written essays is much higher than that of AI-generated essays, meaning human writing is less predictable to language models. Humans often use unexpected phrasing, varied structures, and idiosyncratic styles, while AI text tends to follow smoother, more statistically likely patterns. Higher perplexity thus reflects the greater variability and originality of human language compared with machine-generated text. Chronically, the newer models (GPT-4o, Gemini) showed slightly higher perplexity than older models, which suggests that the essays generated by these models deviated further from the GPT-2 baseline model than the other LLMs. The perplexity of essays from Llama shows the largest variability, indicating that the essays may not be consistently generated. These perplexity values serve as useful reference points for identifying essays generated by different LLMs.

\subsection{Comparing Scores}
\begin{table*}[th!]
\centering
\newcommand*\rot[1]{\rotatebox{90}{#1}}
\footnotesize
\setlength{\tabcolsep}{2pt}
\begin{tabular}{cc|ccc|cccccccc}

\toprule
\textbf{Type} & 	\textbf{Scores} & 	\textbf{GPT-4o} &	\textbf{Gemini}  & \textbf{Llama3-8b}  & \textbf{GPT-4} &\textbf{GPT-3.5}  & \textbf{Bard}  &	\textbf{Mistral} & \textbf{Koala} & \textbf{Vicuna}  & \textbf{Llama}\\
& & \multicolumn{3}{c|}{(mid-2024)} & \multicolumn{7}{c}{(mid-2023)} \\
\midrule
& & \faLock & \faLock & \faUnlock & \faLock & \faLock & \faLock & \faUnlock & \faUnlock & \faUnlock & \faUnlock  \\
\midrule
\multirow{8}{*}{\rot{\textbf{Human Rater}}}& 6  & 6.5\% & 6.5\% & 0.5\% & 2.5\% & 4.0\%  & 0.5\% & 0.5\% & 0.5\% & 0.0\% & 0.0\% \\
& 5  & 54.5\% & 64.5\% & 16.5\% & 48.0\% & 36.5\%  & 12.5\% & 5.0\% & 2.5\% & 8.0\% & 0.0\% \\
& 4  & 38.5\% & 29.0\% & 82.5\%  & 49.5\% & 57.0\% & 75.5\% & 73.5\% & 72.5\% & 71.5\% & 1.5\% \\
& 3  & 0.5\% & 0.0\% & 0.5\% & 0.0\% & 2.5\%   & 11.5\% & 21.0\% & 24.0\% & 20.5\% & 20.5\% \\
& 2  & 0.0\% & 0.0\% & 0.0\% & 0.0\% & 0.0\% & 0.0\% & 0.0\% & 0.5\% & 0.0\% & 41.5\% \\
& 1  & 0.0\% & 0.0\% & 0.0\% & 0.0\% & 0.0\% & 0.0\% & 0.0\% & 0.0\% & 0.0\% & 32.0\% \\
& 0  & 0.0\% & 0.0\% & 0.0\% & 0.0\% & 0.0\% & 0.0\% & 0.0\% & 0.0\% & 0.0\% & 4.5\% \\
\midrule
& Mean  & \underline{4.67} & \textbf{4.78} & 4.17 & 4.53 & 4.42 & 4.02 & 3.85 & 3.79 & 3.88 & 1.82 \\
\midrule
\midrule
\multirow{8}{*}{\rot{\textbf{e-rater\textsuperscript{®}}}}& 6  & 31.5\% & 7.5\% & 1.5\% & 0.0\% & 3.5\% & 1.5\%  & 1.0\% & 1.0\% & 0.5\% & 1.0\% \\
& 5  & 68.5\% & 92.5\% & 93.0\% & 98.0\% & 93.5\%  & 71.5\% & 52.5\% & 43.0\% & 54.0\% & 33.5\% \\
& 4  & 0.0\% & 0.0\% & 5.5\% & 0.5\% & 3.0\% & 28.5\% & 44.5\% & 55.5\% & 45.5\% & 44.0\% \\
& 3  & 0.0\% & 0.0\% & 0.0\% & 0.0\% & 0.0\% & 0.0\% & 2.0\% & 0.5\% & 0.0\% & 14.5\% \\
& 2  & 0.0\% & 0.0\% & 0.0\% & 0.0\% & 0.0\% & 0.0\% & 0.0\% & 0.0\% & 0.0\% & 5.0\% \\
& 1  & 0.0\% & 0.0\% & 0.0\% & 0.0\% & 0.0\% & 0.0\% & 0.0\% & 0.0\% & 0.0\% & 1.0\% \\
& 0  & 0.0\% & 0.0\% & 0.0\% & 0.0\% & 0.0\% & 0.0\% & 0.0\% & 0.0\% & 0.0\% & 1.0\% \\
\midrule
& Mean  & \textbf{5.32} & \underline{5.08} & 4.96 & 5.01 & 5.00 & 4.71 & 4.53 & 4.45 & 4.55 & 4.04 \\

\bottomrule
\end{tabular}
\caption{The percentage of essays in different score levels based on human raters and e-rater\textsuperscript{®}. The symbols \faUnlock~ and \faLock~ indicate open source and proprietary LLMs, respectively. The performance of the best and second-best models, as measured by the average of different scoring systems, is indicated in \textbf{bold} and \underline{underlined}, respectively. }\label{tab:human_erator_score}
\end{table*}

Table \ref{tab:human_erator_score} presents the score distribution. For scores assigned by human raters, essays generated by proprietary LLMs, such as GPT models and Gemini, generally received higher scores than those produced by open-source models. Additionally, essays from newer models released in 2024 tended to receive higher scores than those from models introduced in 2023. The average scores of essays from top LLMs are between 4 and 5, which correspond to ``presents a competent analysis of the issue and conveys meaning with acceptable clarity'' and ``generally thoughtful, well-developed analysis of the issue and conveys meaning clearly,'' respectively, according to the GRE scoring framework. Compared to human raters, the e-rater\textsuperscript{®} engine consistently assigned higher scores to AI-generated essays, and the essays from proprietary models also get higher e-rater\textsuperscript{®} scores compared to those from open-source models. The inflated e-rater\textsuperscript{®} score reveals important limitations in using the current system to evaluate essays generated (or significantly assisted) by AI. 

As noted earlier, the e-rater score is determined by two parts: the language features and their weights, which are calibrated based on human-written essays. Since LLMs generate text based on learned probabilities from large language corpora, they tend to perform systematically better on many language features, such as grammar, mechanics, and usage. When e-rater\textsuperscript{®} continues to apply the same feature weights originally optimized for human-written essays, it produces systematically higher scores for AI-generated content. This is not necessarily a flaw, given that it simply reflects the fact that AI-generated essays are doing much better in these aspects. 

On the other hand, the language features used by e-rater\textsuperscript{®} have traditionally served as reliable indicators of writing quality in human-authored essays, as strong language use often correlates with coherent reasoning and meaningful content. However, this relationship weakens with AI-generated texts, which appear fluent and with fewer errors in grammar, mechanics, and usage but often lack substance and depth. As a result, the existing set of language features used by e-rater does not cover deeper aspects of writing, such as logical and analytical reasoning, which are closely related to specific contexts. Without features that reflect these qualities, e-rater\textsuperscript{®} is unable to evaluate the full depth of AI-generated writing. In contrast, experienced human raters can recognize the issues in AI-generated essays and adjust their scores accordingly. This mismatch in evaluative capability contributes to the observed discrepancy in scores between human raters and e-rater.

This discrepancy in scores between human raters and e-rater underscores the need for and possible directions to improve e-rater (or similar automated scoring engines) to better handle essays in the AI era, where AI-generated and AI-assisted writing is increasingly common. The improvements can be made in two areas: developing new features that cover deeper levels of thinking in the essay and recalibrating the feature weights to contribute to the final score. We want to emphasize that this does not necessarily mean the current version of e-rater\textsuperscript{®} is flawed for its intended use for scoring human-written essays. It remains valid and reliable in that setting, where the existing language features are generally well aligned with writing quality. Its limitations arise only when the system is applied to texts created or influenced by LLMs, a scenario likely to become increasingly prevalent in the near future.

\subsection{Comparing the Detection of AI-generated Essays}

To address \textbf{RQ3}, we compare the performance of the detection of AI-generated essays in two different scenarios. For the within-model scenario, where the detectors are trained and tested using essays generated from the same LLM, Table \ref{tab:detector_1} shows that, overall, the classifiers based on perplexity features slightly outperformed those using the e-rater language features. Notable exceptions include the GPT-4o and Gemini models, where language features are slightly more effective. These findings suggest that when the LLM is known, essays generated by that model can be accurately detected in the context of GRE analytical writing. It is important to note that these detection benchmarks reflect only the detection of AI-generated essays without human revisions. If AI-generated text is revised by humans or mixed with human-written content, detection accuracy is expected to decrease \citep{haoandfauss, hao2024transforming}.

\begin{table*}[th!]
\centering
\footnotesize

\begin{tabular}{llrrrrr}
\toprule
Feature Set & LLM & Accuracy & Precision & Recall & F1 Score & ROC AUC \\
\midrule
Language & \multirow{2}{*}{GPT-4o} & 0.988 & 0.990 & 0.985 & 0.988 & 1.000 \\
Perplexity & & 0.970 & 0.970 & 0.970 & 0.970 & 0.995 \\ 
\midrule
Language & \multirow{2}{*}{Gemini}  & 0.995 & 0.995 & 0.995 & 0.995 & 0.997 \\
Perplexity &  & 0.910 & 0.939 & 0.880 & 0.908 & 0.960 \\
\midrule
Language & \multirow{2}{*}{Llama3-8b}  & 0.972 & 0.970 & 0.975 & 0.972 & 0.998 \\
Perplexity&  & 0.997 & 0.995 & 1.000 & 0.998 & 0.997 \\
\midrule
Language &  \multirow{2}{*}{GPT-4} & 0.975 & 0.976 & 0.975 & 0.975 & 0.997 \\
Perplexity & & 0.997 & 0.995 & 1.000 & 0.998 & 1.000 \\
\midrule
Language & \multirow{2}{*}{GPT-3.5} & 0.957 & 0.956 & 0.960 & 0.957 & 0.995 \\
Perplexity & &0.997 & 1.000 & 0.995 & 0.997 & 0.997 \\
\midrule
Language & \multirow{2}{*}{Bard} & 0.960 & 0.958 & 0.965 & 0.960 & 0.997 \\
Perplexity &  & 0.997 & 1.000 & 0.995 & 0.997 & 0.997 \\
\midrule
Language & \multirow{2}{*}{Mistral} & 0.963 & 0.961 & 0.965 & 0.963 & 0.996 \\
Perplexity &  & 0.993 & 0.995 & 0.990 & 0.992 & 1.000 \\
\midrule
Language & \multirow{2}{*}{Koala}  & 0.990 & 0.986 & 0.995 & 0.990 & 0.999 \\
perplexity &  & 0.997 & 0.995 & 1.000 & 0.998 & 0.997 \\
\midrule
Language & \multirow{2}{*}{Vicuna}  & 0.985 & 0.976 & 0.995 & 0.985 & 1.000 \\
Perplexity &  & 0.997 & 0.995 & 1.000 & 0.998 & 0.997 \\
\midrule
Language& \multirow{2}{*}{Llama} & 0.948 & 0.955 & 0.940 & 0.947 & 0.988 \\
Perplexity  &  & 0.997 & 1.000 & 0.995 & 0.997 & 0.997 \\
\bottomrule
\end{tabular}

\caption{Performance of the XGBoost classifier with different feature sets: e-rater\textsuperscript{®} language features, perplexity feature 
 on the binary classification task. \textit{Note: Results are based on the average of five-fold cross-validation.}}\label{tab:detector_1}
\end{table*}

\begin{figure}
    \centering
    \includegraphics[width=0.5\linewidth]{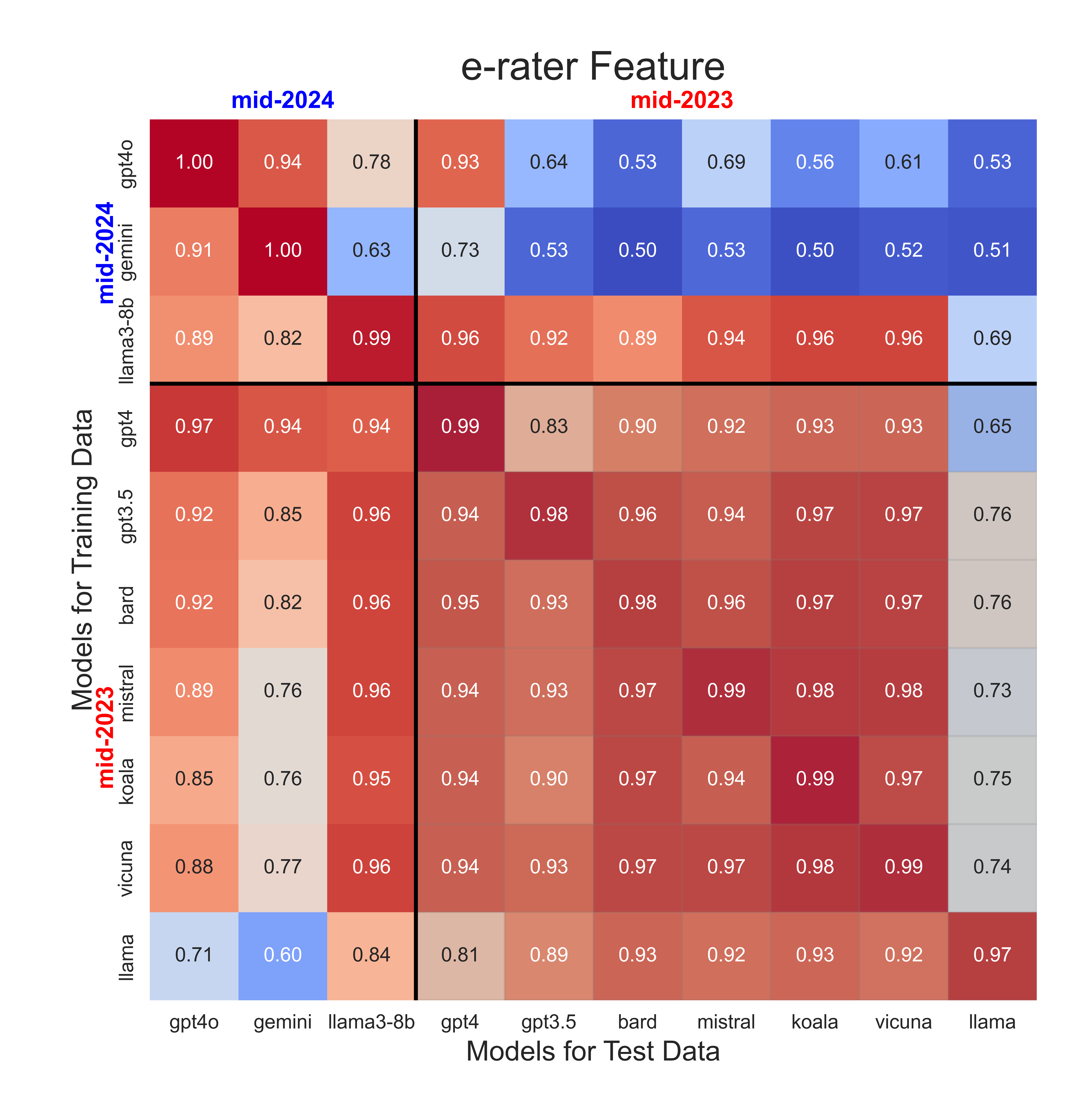}
    \caption{Detection accuracy for classifiers trained on e-rater\textsuperscript{®} language features}
    \label{fig:erater-pair}
\end{figure}

\begin{figure}
    \centering
    \includegraphics[width=0.5\linewidth]{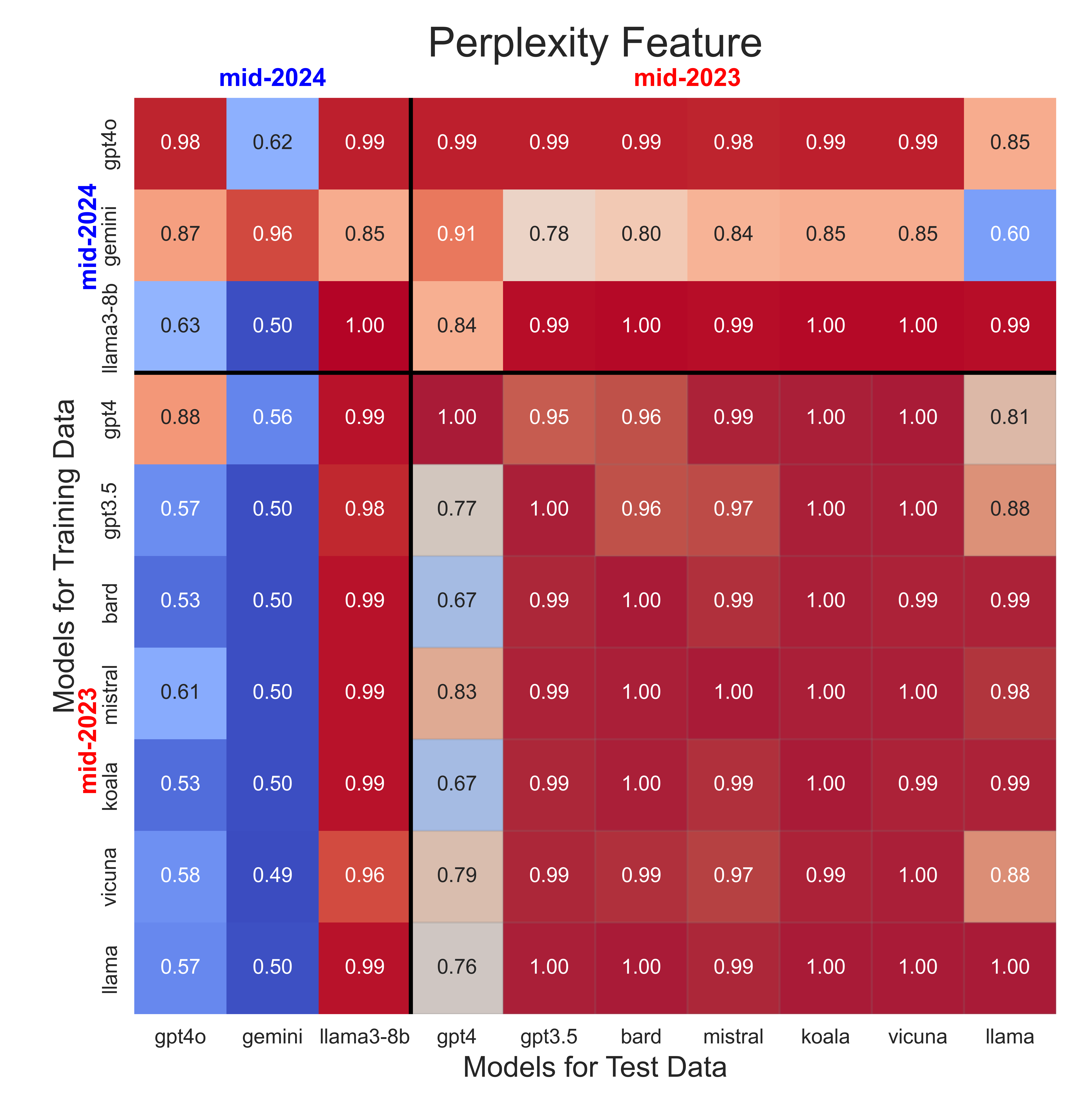}
    \caption{Detection accuracy for classifiers trained on perplexity-based features}
    \label{fig:perplexity-pair}
\end{figure}

For the cross-model scenario, where the classifier is trained using essays generated by one LLM and tested on essays generated by a different LLM, Figure \ref{fig:erater-pair} and \ref{fig:perplexity-pair} how the results based on five-fold cross-validation. Overall, perplexity feature-based classifiers performed slightly better than language feature-based models when the training and testing data came from LLMs of the same year. They also excelled when trained on data from 2024 LLMs and tested on 2023 AI-generated data. On the other hand, language feature-based classifiers performed significantly better when trained on data from 2023 LLMs and tested on data from 2024 LLMs. Surprisingly, classifiers trained on GPT-4o data using perplexity-based features achieved very high accuracy in detecting essays from most LLMs, with the exception of essays generated by Gemini and the 2023 version of Llama. This suggests that perplexity features are generally effective across models but may struggle with certain newer or more distinct language patterns in specific LLMs.

\section{Discussion}

Over the past three years, especially since the release of ChatGPT, LLMs and generative AI have advanced at an unprecedented pace. Increasingly powerful LLMs are now widely used in various writing tasks and have been integrated into commonly used tools such as Microsoft Word and Google Docs. This widespread adoption signals the beginning of an era of co-created writing, where human and AI contributions are increasingly intertwined. This shift raises fundamental questions for writing assessment, including what aspect of writing should be assessed, how it should be assessed, and how academic integrity can be maintained in an AI-mediated writing environment.

In this paper, we conducted a large-scale empirical study of AI-generated essays through the lens of the GRE Analytical Writing assessment to deepen understanding of their characteristics and examine their implications for two critical components of writing assessment: automated scoring and academic integrity. We compared the characteristics of essays generated by 10 popular LLMs and humans and observed that proprietary LLMs, such as GPTs and Gemini, consistently produced higher-scoring essays compared to open-source models, aligning with existing evaluations that have shown proprietary models outperforming their open-source counterparts. Additionally, we observed a notable improvement in essay scores from 2024 models compared to 2023 models, reflecting the continuous efforts within the AI community to create increasingly capable models for generating higher-quality text. Further analysis of the language features computed by the e-rater\textsuperscript{®} engine revealed that AI-generated essays better controlled grammar, usage, and mechanics errors than human-written essay baseline. All AI-generated essays exhibit a lower perplexity than the baseline of human-written essays, which substantiates that perplexity is a crucial indicator for detecting AI-generated essays across LLMs. These findings provide valuable insights for improving future AI text generation, emphasizing the crafting of more meaningful texts alongside their already high-quality linguistic characteristics.

Another notable observation is that e-rater\textsuperscript{®} systematically assigns higher scores to AI-generated essays compared to human raters. This discrepancy highlights limitations in current automated scoring systems when applied to AI-generated or AI-assisted writing. It also suggests directions for future improvements, such as developing new features that capture deeper meaning and reasoning and recalibrating the weighting of existing language features to more accurately reflect their contribution when AI tools can effectively handle them.

Additionally, we examined the accuracy of machine learning classifiers that use language features from e-rater\textsuperscript{®} and perplexity-based features to distinguish between human-written and AI-generated essays, considering both within-model and cross-model scenarios. The classifiers demonstrated strong performance, indicating that essays generated by AI in response to GRE Analytical Writing prompts can be reliably identified. It is important to note, however, that these results apply specifically to essays that are entirely generated by AI. When human edits are introduced or when the writing involves more iterative human-AI collaboration, detection becomes more challenging. Nonetheless, these findings provide a useful benchmark for detection accuracy in the context of GRE Analytical Writing, and the results are consistent with findings from other writing assessments \citep{haoandfauss, Yan2023-ha}
.

In summary, this study empirically examined essays generated by several major LLMs in the context of GRE Analytical Writing, benchmarked their characteristics, and analyzed the implications of AI-generated or assisted essays for automated scoring and academic integrity in the AI era. These findings aim to inform the assessment community about the current capabilities of AI-generated writing and its impact on writing evaluation practices. In the meantime, we hope the findings can also help the LLM community think about ways to enhance the quality of model outputs through improved training and fine-tuning techniques.

\section*{Limitations}
This study has several limitations. First, LLMs are rapidly evolving, making it challenging to capture the full range of advanced models available at any given time. As a result, the characteristics and quality of AI-generated essays are subject to continuous change. Ongoing research will be necessary to monitor how the findings presented here may shift as newer models are developed and adopted.

Secondly, it is crucial to note that the quality of generated texts heavily hinges on the provided prompts. In our study, we used the original prompts from the analytical writing assessment, only adding a length requirement. However, we acknowledge that modifying prompts in several ways, such as instructing the model to write as if it were a six-year-old, can dramatically alter the quality of the generated essay. As there are infinite ways to tweak the prompts, our choice is motivated by the idea of modifying the prompts as little as possible to simplify and obtain baseline results for benchmarking purposes. In future work, we will explore the impact of prompt tweaking on the essay quality. 

Third, the variability of the generated essay is controlled by hyper-parameters, such as the temperature of the LLMs. Different LLMs may have different baseline settings for these hyper-parameters, which leads to varying degrees of variability in generated essays, even when using the same hyperparameter values. Therefore, a systematic investigation of the quality of essays for different hyper-parameters is imperative, and we will report the findings in future work.

Finally, we proposed several high-level strategies for improving automated scoring systems when evaluating AI-generated or AI-assisted essays. However, we did not provide detailed specifications for new features or weight adjustment methods. This is partly because the quality of AI-generated essays is highly sensitive to the specific language model and prompting strategy employed. Additionally, our suggestions are based on a specific scoring engine, e-rater\textsuperscript{®}, which may not fully reflect the challenges faced by other automated scoring systems. These issues warrant careful follow-up, and we plan to report further findings in future work.

\bibliography{references}

\appendix

\section{GRE Scoring Framework}\label{appendix:complete_table1}
The scoring framework of GRE “analyze an issue” writing task (Retrieved from GRE General Test Analytical Writing Scoring (ets.org))

\noindent\textbf{Score 6: Outstanding}\\
A 6 response presents a cogent, well-articulated analysis of the issue and conveys meaning skillfully. A typical response in this category:
\begin{itemize}
    \item Articulates a clear and insightful position on the issue in accordance with the assigned task.
    \item Develops the position fully with compelling reasons and/or persuasive examples.
    \item Sustains a well-focused, well-organized analysis, connecting ideas logically.
    \item Conveys ideas fluently and precisely, using effective vocabulary and sentence variety.
    \item Demonstrates superior facility with the conventions of standard written English (i.e., grammar, usage, and mechanics), but may have minor errors
\end{itemize}

\noindent\textbf{Score 5: Strong}\\
A 5 response presents a generally thoughtful, well-developed analysis of the issue and conveys meaning clearly. A typical response in this category:

\begin{itemize}
    \item Presents a clear and well-considered position on the issue in accordance with the assigned task.
    \item Develops the position with logically sound reasons and/or well-chosen examples.
    \item Is focused and generally well organized, connecting ideas appropriately.
    \item Conveys ideas clearly and well, using appropriate vocabulary and sentence variety.
    \item Demonstrates facility with the conventions of standard written English but may have minor errors
\end{itemize}

\noindent\textbf{Score 4: Adequate}\\
A 4 response presents a competent analysis of the issue and conveys meaning with acceptable clarity. A typical response in this category:

\begin{itemize}
\item Presents a clear position on the issue in accordance with the assigned task.
\item Develops the position with relevant reasons and/or examples.
\item Is adequately focused and organized.
\item Demonstrates sufficient control of language to express ideas with acceptable clarity.
\item Generally demonstrates control of the conventions of standard written English, but may have some errors
\end{itemize}

\noindent\textbf{Score 3: Limited}\\
A 3 response demonstrates some competence in addressing the specific task directions, in analyzing the issue and in conveying meaning, but is obviously flawed. A typical response in this category exhibits one or more of the following characteristics:

\begin{itemize}
\item Is vague or limited in addressing the specific task directions and in presenting or developing a position on the issue or both.
\item Is weak in the use of relevant reasons or examples or relies largely on unsupported claims.
\item It is limited in focus and/or organization.
\item Has problems in language and sentence structure that result in a lack of clarity.
\item Contains occasional major errors or frequent minor errors in grammar, usage or mechanics that can interfere with meaning
\end{itemize}

\noindent\textbf{Score 2: Seriously Flawed}\\
A 2 response largely disregards the specific task directions and/or demonstrates serious weaknesses in analytical writing. A typical response in this category exhibits one or more of the following characteristics:

\begin{itemize}
\item Is unclear or seriously limited in addressing the specific task directions and in presenting or developing a position on the issue or both.
\item Provides few, if any, relevant reasons or examples in support of its claims
\item Is poorly focused and/or poorly organized.
\item Has serious problems in language and sentence structure that frequently interfere with meaning.
\item Contains serious errors in grammar, usage or mechanics that frequently obscure meaning
\end{itemize}

\noindent\textbf{Score 1: Fundamentally Deficient}\\
A 1 response demonstrates fundamental deficiencies in analytical writing.
A typical response in this category exhibits one or more of the following characteristics:

\begin{itemize}
\item Provides little or no evidence of understanding the issue.
\item Provides little or no evidence of the ability to develop an organized response (e.g., is disorganized and/or extremely brief)
\item Has severe problems in language and sentence structure that persistently interfere with meaning.
\item Contains pervasive errors in grammar, usage or mechanics that result in incoherence
\end{itemize}

\noindent \textbf{Score 0: Fundamentally Deficient}\\
Off topic (i.e., provides no evidence of an attempt to address the assigned topic), is in a foreign language, merely copies the topic, consists of only keystroke characters or is illegible or nonverbal.

\section{LLM Model Details}\label{appendix:llm_detail}

We listed the details of LLMs use used for essay generation in Table \ref{tab:llm_date}.
\begin{table*}[]
    \centering
    \footnotesize
    \begin{tabular}{l|c|c}
    \toprule
    \textbf{LLM} & \textbf{Version} & \textbf{URL}\\
    \midrule
    GPT-4o    & 2024-05-13 & \url{https://openai.com/index/hello-gpt-4o/}\\
       Gemini (1.5 Pro) & 2024-05-14 & \url{https://gemini.google.com} \\
       Llama3-8B-Instruct & 2024-04-18 & \url{https://huggingface.co/meta-llama/Meta-Llama-3-8B-Instruct} \\
       \midrule
       GPT-4    & 2024-05-13 & \url{https://platform.openai.com/docs/models}\\
       GPT3.5-turbo & 2023-06-13 & \url{https://platform.openai.com/docs/models} \\
       Bard & 2023-06-07 &  (now upgraded to Gemini) \\
       Mistral & 2023-09-27 & \url{https://mistral.ai/news/announcing-mistral-7b/}\\
       Vicuna-13B & 2023-03-30 & \url{https://lmsys.org/blog/2023-03-30-vicuna/}\\
       Koala-13B & 2023-04-03 & \url{https://bair.berkeley.edu/blog/2023/04/03/koala/}\\
       Llama1-13B & 2023-02-24 & \url{https://ai.meta.com/blog/large-language-model-llama-meta-ai/}\\
       
       \midrule
       
    \end{tabular}
    \caption{LLMs used for our experiments.}
    \label{tab:llm_date}
\end{table*}

\end{document}